\documentclass[runningheads]{llncs}

\usepackage{accv}

\usepackage{accvabbrv}

\usepackage{graphicx}
\usepackage{booktabs}
\usepackage{multirow}
\usepackage{colortbl}
\usepackage{algorithm}
\usepackage[dvipsnames]{xcolor}
\usepackage{algpseudocode}
\usepackage{wrapfig}

\definecolor{lightgray}{rgb}{0.95, 0.95, 0.95}

\usepackage[accsupp]{axessibility}  

\usepackage{hyperref}

\usepackage{orcidlink}

\begin{document}

\title{GTA: Global Tracklet Association for Multi-Object Tracking in Sports} 


\author{Jiacheng Sun\inst{1}\orcidlink{0009-0009-8710-7858} \and
Hsiang-Wei Huang\inst{1}\orcidlink{0009-0009-2474-8869} \and
Cheng-Yen Yang\inst{1}\orcidlink{0009-0004-2631-6756} \and
Zhongyu Jiang\inst{1}\orcidlink{0000-0003-4462-6497} \and
Jenq-Neng Hwang\inst{1}\orcidlink{0000-0002-8877-2421}}
\institute{University of Washington, Seattle WA, USA\inst{1}\\
\email{\{sjc042, hwhuang, cycyang, zyjiang, hwang\}@uw.edu}}

\authorrunning{J. Sun et al.}

\maketitle
\begin{abstract}

Multi-object tracking in sports scenarios has become one of the focal points in computer vision, experiencing significant advancements through the integration of deep learning techniques. Despite these breakthroughs, challenges remain, such as accurately re-identifying players upon re-entry into the scene and minimizing ID switches. In this paper, we propose an appearance-based global tracklet association algorithm designed to enhance tracking performance by splitting tracklets containing multiple identities and connecting tracklets seemingly from the same identity. This method can serve as a plug-and-play refinement tool for any multi-object tracker to further boost their performance. The proposed method achieved a new state-of-the-art performance on the SportsMOT dataset with HOTA score of 81.04\%. Similarly, on the SoccerNet dataset, our method enhanced multiple trackers' performance, consistently increasing the HOTA score from 79.41\% to 83.11\%. These significant and consistent improvements across different trackers and datasets underscore our proposed method's potential impact on the application of sports player tracking. We open-source our project codebase at \href{https://github.com/sjc042/gta-link.git}{https://github.com/sjc042/gta-link.git}.
\end{abstract}

\section{Introduction}
\label{sec:intro}

In recent years, advancements in computer vision and deep learning have revolutionized sports analytics, offering unprecedented insights into player performance and strategy. For example, sports video understanding~\cite{Cioppa2020Context, Duan_2022}, sports field registration~\cite{jiang2020optimizing, Gutierrez-Perez_2024_CVPR}, and 2D/3D human pose estimation for sports~\cite{jiang2022golfpose}. Among these innovations, sports player tracking systems have emerged as a cornerstone, providing coaches and analysts with valuable data on player movements, positioning, and interactions during game play \cite{Gade_2020_CVPR}. These systems have become integral to modern sports, enabling data-driven decision-making and performance optimization across various disciplines.

However, despite significant progress in multi-object tracking technologies, challenges persist in accurately tracking players, including the irregular movements and similar appearances of sports players, and the lack of re-identification algorithm in handling re-entry situation after leaving camera field of view after certain amount of time.

Current state-of-the-art on-line tracking algorithms often fail in long-term object re-identification with the aforementioned challenges. While these trackers perform well in controlled environments or short-term scenarios, they struggle to maintain consistent player identities throughout entire matches or when players re-enter the field after substantial absences. This limitation significantly impacts the accuracy and reliability of player performance analysis, tactical evaluations, and automated game statistics.

\begin{figure}[t]
\centering
\includegraphics[width=1\linewidth]{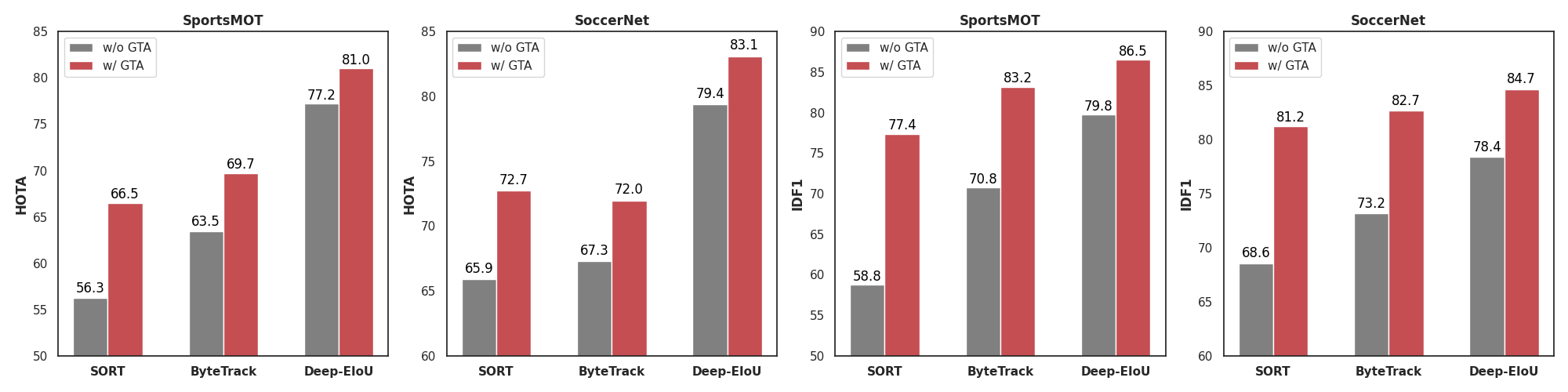}
\vspace{-1em}
\caption{Our proposed Global Tracklet Association (GTA) method significantly boosts the HOTA and IDF1 score of existing trackers, such as SORT, ByteTrack, and Deep-EIoU, on sports tracking datasets, including SportsMOT and SoccerNet.}
\label{SportsPerformance}
\end{figure}

To address these persistent issues, our paper proposes an effective plug-and-play post-processing algorithm named Global Tracklet Association (GTA), designed specifically for sports player tracking applications. GTA aims to refine the tracking results of on-line or off-line trackers by improving long-term re-identification capabilities and handling the unique challenges posed by sports environments. By leveraging global temporal information and advanced association techniques, GTA enhances the robustness and accuracy of player tracking, potentially bridging the gap between current tracking technologies and the demanding requirements of professional sports analytics.
\section{Related Work}

\begin{wrapfigure}{r}{0.5\textwidth}
\vspace{-6em}
\centering
\includegraphics[width=0.8\linewidth]{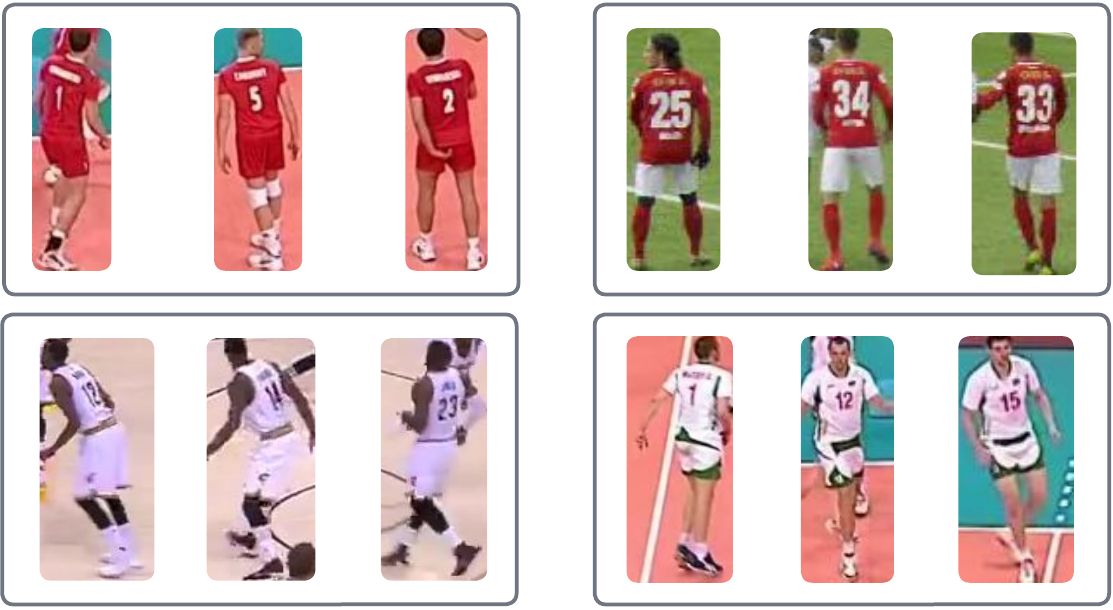}
\caption{\footnotesize{Examples of \textbf{different players} on the same teams, highlighting the challenge of distinguishing between players with similar appearances in sports tracking.}}
\vspace{-2em}
\label{similars}
\end{wrapfigure}

\subsection{Sports Player Tracking}

With progress made in object detection and tracking, recent studies have focused on challenging multi-object tracking (MOT) scenarios like sports \cite{cui2023sportsmot, cioppa2022soccernet} and dancing \cite{sun2022dancetrack}. Tracking sports players is more difficult than tracking pedestrians due to the complex nature of sports scenarios, including frequent occlusions, rapid direction changes, varying player densities, and similar appearance as in Figure \ref{similars}, and re-entries to the camera view. Several works \cite{huang2024iterative, OCSORT, huang2023observation, yang2023hard} have proposed methods to handle irregular object motion, improving tracking performance compared to traditional Kalman filter-based methods \cite{SORT}. 

\begin{figure}[t]
\centering
\includegraphics[width=0.85\linewidth]{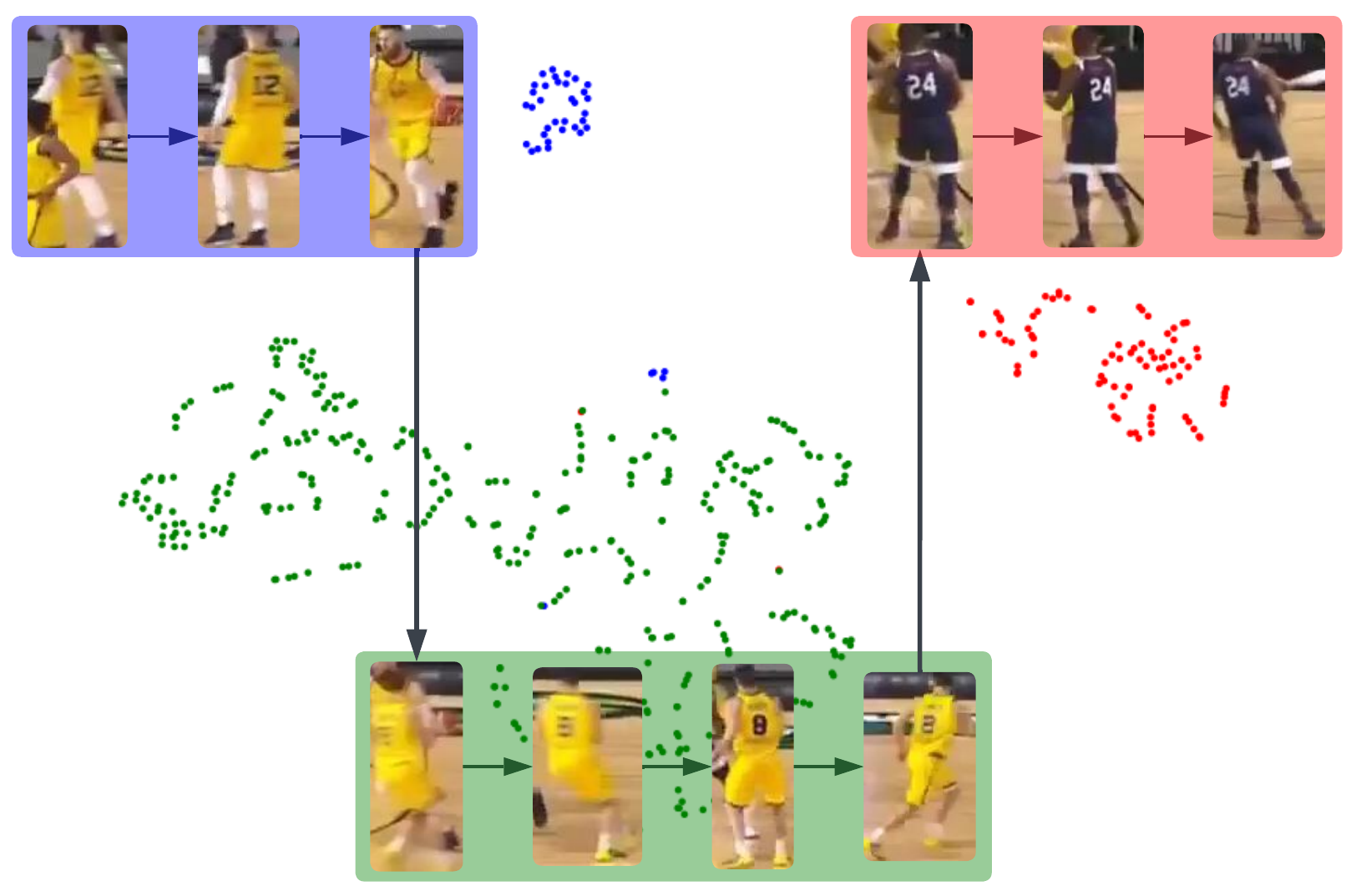}
\caption{An example of a mix-up error in a single tracklet. The tracklet output by the online tracking system contains \textbf{three} different identities, represented by purple, green, and red points. The figure illustrates the tracklet's features extracted by a ReID model and clustered using the DBSCAN clustering algorithm.}
\label{fig:mix-up-error}
\end{figure}

\begin{figure}[t]
\centering
\vspace{3em}
\includegraphics[width=1\linewidth]{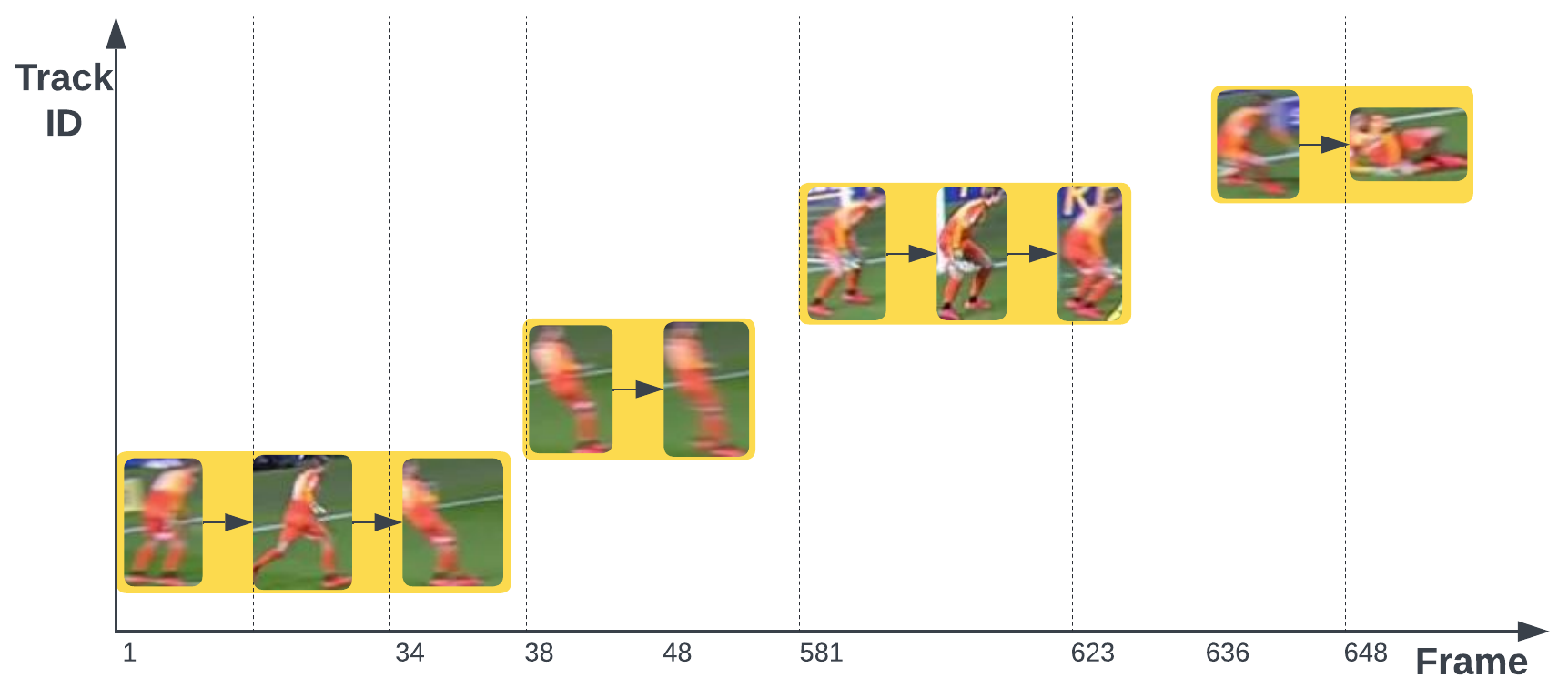}
\caption{An example of a cut-off error, where a player's tracklet is fragmented into four separate segments due to the player exiting and re-entering the camera view multiple times throughout the video sequence.}
\label{fig:cut-off-error}
\end{figure}

There are two predominant types of errors in sports player tracking: the mix-up error (Figure~\ref{fig:mix-up-error}) and the cut-off error (Figure~\ref{fig:cut-off-error}). The first type, mix-up errors, occur due to irregular movements and occlusions during tracking, leading to a single tracklet mistakenly including multiple players (Figure~\ref{fig:mix-up-error}). The second type, cut-off errors, arise because, unlike targets in traditional pedestrian tracking datasets \cite{mot16}, sports players often re-enter the camera's view after exiting. Assigning a new tracking ID to a previously tracked player results in a cut-off error, which fragments a single player's tracklet into multiple parts (Figure~\ref{fig:cut-off-error}).

\subsection{Person Re-Identification}
When a player re-appears in the scene, a re-identification method is needed to assign the correct tracking ID. Although some trackers can re-identify players who re-enter shortly after exiting by lengthening the tracking buffer, cut-off errors from extended absences and re-entries are most prevalent in sports tracking. Person re-identification (ReID) is crucial in multi-object tracking for identifying individuals across video instances or different cameras.

Several methods have been proposed to address ReID challenges. OSNet \cite{OSNet} introduced a lightweight CNN-based method with state-of-the-art performance on various ReID datasets. Many tracking methods incorporate ReID models in the data association stage, such as DeepSORT \cite{DeepSORT} and BoTSORT \cite{aharon2022bot}. In sports multi-player tracking, Deep-EIoU \cite{huang2024iterative} has shown that using a ReID model can significantly improve tracking performance in sports player tracking scenarios.

\subsection{Global Link Models}
To address cut-off errors, where a tracker incorrectly assigns different tracking IDs to the same target, several previous works have proposed global linking models that utilize various types of information, such as appearance, spatial, and temporal cues, to associate fragmented tracklets and reassign the correct tracking IDs after the online tracking process. These models either utilize motion or appearance features of tracklets for tracklet-level association. For example, Translink \cite{zhang2023translink} incorporates a CNN and temporal attention network to extract and encode a tracklet's appearance features, treating the merging process of tracklet pairs as a binary classification task. AFLink \cite{du2023strongsort} uses only spatial-temporal information. Some methods \cite{huang2023enhancing, yang2024online, cherdchusakulchai2024online} utilize feature clustering methods to merge tracklets and boost the performance on multi-camera tracking scenarios. MambaTrack \cite{huang2024exploring} proposed a motion model that serves as a motion predictor and extracts tracklet motion features for further global tracklet association.

Additionally, some methods exploit object moving direction \cite{huang2023multi,hsu2019multi} or metadata \cite{yang2023sea} as clues to conduct global tracklet association and enhance tracking performance. \cite{wang2022split} proposed a universal tracklet booster based on CNN and temporal attention to address both mix-up and cut-off errors.

In this work, we propose a novel plug-and-play box-grained global tracklet association model, including a tracklet splitter and a connector. The proposed method is specifically designed to conduct player re-identification and boost the tracking performance in various sports scenarios.

\section{Methods}
Drawing inspiration from global link models presented in recent works \cite{huang2023observation} \cite{yang2023hard}, we propose the Global Tracklet Association (GTA) method, a novel plug-and-play approach designed to address both mix-up and cut-off errors in multi-object tracking for sports scenarios. Our method consists of two key modules: a \textbf{\textit{Tracklet Splitter}} and a \textbf{\textit{Tracklet Connector}}, which leverage deep feature representations and spatial constraints to enhance tracking accuracy and robustness.

\begin{figure}[t]
\centering
\includegraphics[width=0.85\linewidth]{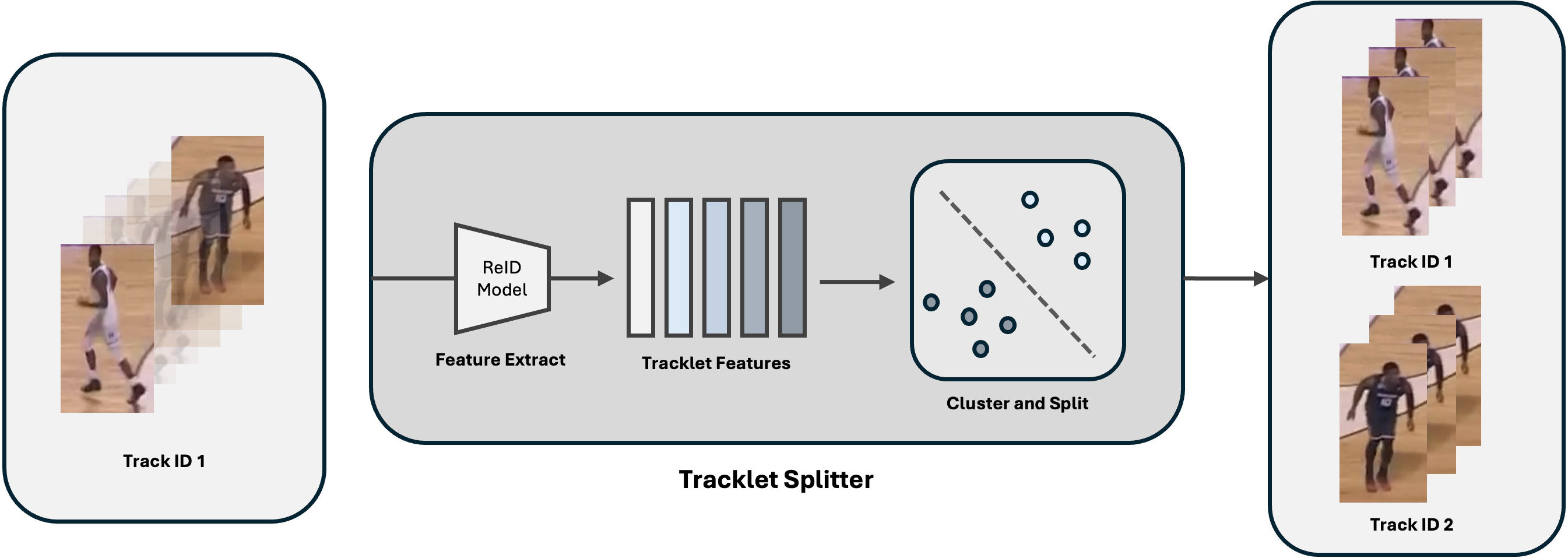}
\caption{Illustration of tracklet splitter.}
\label{fig:splitter_pipeline}
\end{figure}

\begin{figure}[t]
\centering
\includegraphics[width=0.85\linewidth]{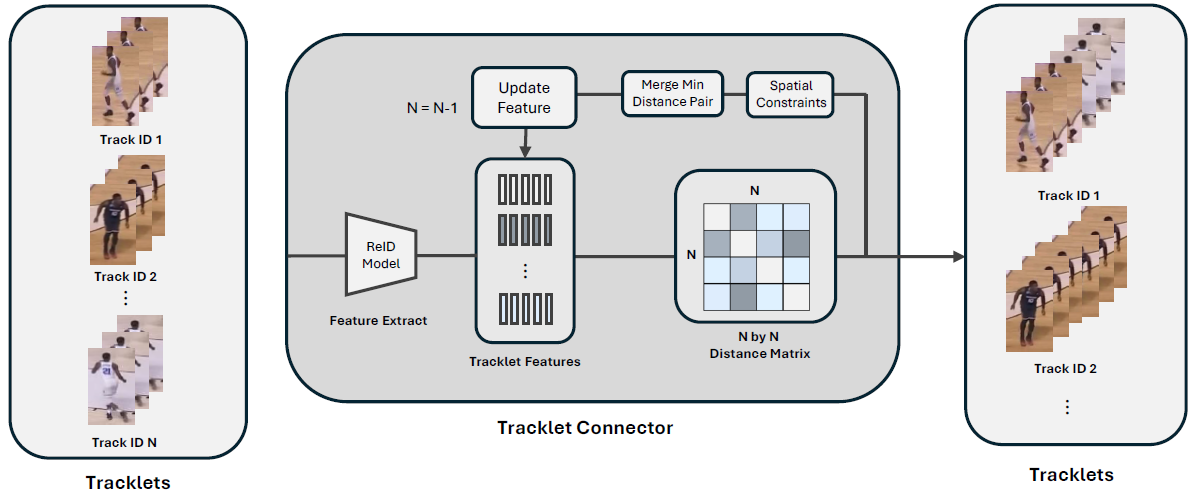}
\caption{Illustration of tracklet connector.}
\label{fig:connector_pipeline}
\end{figure}

Our proposed post-processing method follows a two-stage process to enhance tracking accuracy. Prior to post-processing, box-grained embedding features from online tracking results are generated by a CNN-based ReID model \cite{OSNet} for each tracklet. In the first stage of our tracklet association model, these tracklets are processed through a \textit{\textbf{Tracklet Splitter}} to address mix-up errors, ensuring that instances of different identities are correctly separated, as illustrated in Figure \ref{fig:splitter_pipeline}. In the second stage, the split tracklets belonging to the same identities are further merged by the proposed \textit{\textbf{Tracklet Connector}} to correct cut-off errors, as depicted in Figure \ref{fig:connector_pipeline}.

\subsection{Tracklet Splitter}
The proposed tracklet splitter addresses mix-up errors within a single tracklet, $T = \{t_0, \dots, t_n\}$, by splitting the tracklet into multiple fragments, $t_i$, ensuring that each fragment contains only bounding boxes with similar appearance features, measured by close cosine distances in the feature embedding space. We employ DBSCAN clustering \cite{ester1996density} to split the tracklet into multiple clusters (tracklet fragments) based on their box-grained appearance embedding feature, which are generated by an OSNet ReID model \cite{OSNet}. The pipeline for our tracklet splitter is illustrated in Figure \ref{fig:splitter_pipeline}.

DBSCAN (Density-Based Spatial Clustering of Applications with Noise) is a powerful clustering algorithm that operates by grouping points that are closely packed together while marking points that lie alone in low-density regions as outliers. It is important to note that, unlike traditional DBSCAN implementations, our adapted version assigns outliers to the nearest clusters at the end of the clustering process. This modification is based on the assumption that each bounding box contains a valid detection instance, ensuring that no potentially valuable data points are discarded. This approach allows for a more comprehensive analysis of the tracklet data, preserving information that might otherwise be lost. When applying DBSCAN to the process of tracklet splitting, we incorporate three crucial hyperparameters:

\noindent \textbf{\textit{Minimum Samples ($s$)}}: The Minimum Samples parameter, denoted as $s$, specifies the minimum number of points required to establish a densed region cluster. This parameter serves two critical purposes in the clustering process. First, it ensures that clusters are only formed when there is a sufficient concentration of points, effectively preventing the creation of clusters with too few instances. Second, it aids in noise reduction by initially labeling points that do not meet this threshold as noise or outliers, especially in the case of sport tracking.

\noindent \textbf{\textit{Maximum Neighbor Distance ($\epsilon$)}}: The Maximum Neighbor Distance, represented by $\epsilon$ (epsilon), determines the radius within which points are considered neighbors. This parameter is fundamental in controlling the density requirement for cluster formation. In our implementation, we utilize cosine similarity as the distance metric between fragment features, offering several advantages in the context of tracklet splitting. It controls the density requirement for cluster formation, balancing the need to capture variations within a single identity while distinguishing between different identities. 

\noindent \textbf{\textit{Maximum Clusters ($k$)}}: Unlike the original algorithm, we introduce a maximum clusters parameter, $k$, to limit the number of clusters and prevent excessive fragmentation of the tracklet. If the total number of final clusters exceeds $k$, clusters are progressively merged until only $k$ clusters remain. This modification ensures that a given tracklet is not split into an excessive number of fragments, maintaining a balance between accuracy and fragmentation.

\subsection{Tracklet Connector}
Our tracklet connector is designed to merge fragmented tracklets from the same identity using a hierarchical clustering approach with a distance threshold \textbf{$\alpha$}, addressing the cut-off errors that occur when players exit and re-enter the field 
of view, or the ID switch during the tracking process. Our approach, as depicted in Figure \ref{fig:connector_pipeline}, consists of tracklet clustering based on tracklets distance while applying temporal and spatial constraints to ensure accurate and reliable tracklet merging, and the three main components for the tracklet connector are listed below:

\noindent \textbf{Constructing Tracklet Feature Distance Matrix.}
The first step of our tracklet connector is the initial construction of a symmetric cosine distance matrix that measures the similarity between all tracklet pairs within a video sequence. The distance matrix is constructed as follows:
\begin{equation}
    D_{i,j} = 
    \begin{cases}
        1,\text{ if } i \neq j \And \Pi_{i}\cap\Pi_{j}\neq\emptyset
        \\
        \frac{1} {N_{i}N_{j}} \sum\limits_{i\in\Pi_{i}}
        \sum\limits_{j\in\Pi_{j}}(1 - \frac{F^{i}_{m} \cdot F^{j}_{n}} {\lVert F^{i}_{m}\rVert\ \lVert F^{j}_{n} \rVert}), \text{ otherwise}
    \end{cases}
\label{eq1}
\end{equation}
where $D_{i,j}$ denotes the tracklet distance between tracklet pair $T_{i}$ and $T_{j}$; $\Pi_{i}$ and $\Pi_{j}$ represent temporal spans of tracklet $T_{i}$ and $T_{j}$; $F^{i}_{m}$ and $F^{j}_{n}$ are tracklet $T_{i}$ and $T_{j}$'s embedding feature at frame $m$ and $n$ respectively; $N_{i}$ and $N_{j}$ are the length of the tracklet. This distance enables a nuanced representation of tracklet similarities based on the rich nature of deep features and temporal clues.

\noindent \textbf{Enforcing Spatial Constraints.}
In a common sports game video where the camera position remains fixed, player movement is constrained by the field boundaries, and players do not exit and re-enter from opposite sides of the field. To reflect this, our method enforces spatial constraints for merging tracklets using:
\begin{align}
\theta_{\text{hor}} &= \beta \Delta_{\text{max,hor}},\\
\theta_{\text{ver}} &= \beta \Delta_{\text{max,ver}},
\end{align}
where $\Delta_{\text{max,hor}}$ and $\Delta_{\text{max,ver}}$ denote the maximum horizontal and vertical distances from all bounding boxes in the current video. The spatial factor, $\beta \in (0, 1]$, sets thresholds of $\theta_{\text{hor}}$ and $\theta_{\text{ver}}$, limiting the association distances threshold between temporally adjacent tracklet's exit and entry points using bounding box center. Then, $D_{i,j}$ will be updated. 
\begin{equation}
    D_{i,j} = 1, \text{ if } \Delta_{i,j,hor} > \theta_{hor} \text{ or } \Delta_{i,j,ver} > \theta_{ver},
\end{equation}
where $\Delta_{i,j,hor}$ and $\Delta_{i,j,ver}$ are the horizontal and vertical distance between the beginning and ending of tracklets $i$ and $j$, respectively. This approach filters out unreasonable associations between tracklets, enhancing the accuracy of the tracklets merging process.

\noindent \textbf{Hierarchical Clustering.} After the distance matrix is obtained using equation \ref{eq1}, we further conduct hierarchical clustering following \cite{hsu2019multi} to merge fragment tracklets. We continuously merge the tracklets until no tracklet pair's distance is larger than the merging threshold $\alpha$.

\section{Experiments}

\subsection{Datasets}

We evaluate our method on two large-scale sports player tracking datasets: SportsMOT \cite{cui2023sportsmot} and SoccerNet \cite{cioppa2022soccernet}. 
These datasets are representative of athlete tracking in team sports scenarios, presenting unique challenges such as players with similar appearances, frequent re-entries into the camera's field of view, and abrupt changes in motion.

\noindent \textbf{SportsMOT} is a multi-object tracking dataset that contains over 240 video sequences spanning three team sports: basketball, football, and volleyball. Each sport presents its unique challenges, such as the fast-paced, close-quarters action of basketball, the wide-field dynamics of football, and the rapid vertical movements in volleyball. The dataset provides a robust foundation for developing and testing tracking algorithms in complex, real-world sports environments and has been widely used in benchmarking MOT for sport tracking.

\noindent \textbf{SoccerNet} focuses exclusively on videos captured from soccer matches, providing a collection of over 100 high-quality video clips extracted from professional games. For our experiments, we utilize the test set from the SoccerNet tracking dataset published in 2023.

\subsection{Implementation details}

\noindent\textbf{Detector.} For the SportsMOT test set, we use YOLOX as the detection model following \cite{huang2024iterative}, and for the SoccerNet 2023 test set, we directly apply oracle detection following others' implementations \cite{huang2024iterative} for fair comparison.

\noindent\textbf{Tracker.} In our work, we test our tracklet refinement method on three trackers: SORT \cite{SORT}, ByteTrack \cite{ByteTrack}, and Deep-EIoU \cite{huang2024iterative}. SORT and ByteTrack are both implemented to track with spatial and motion cues. Deep-EIoU incorporates both spatial and appearance information to achieve state-of-the-art performance with a HOTA score of 77.2\% on the SportsMOT test set and 85.4\% on the SoccerNet test set published in 2022.

\noindent\textbf{ReID Model.} For our experiments, we use the OSNet \cite{OSNet} model trained on SportsMOT dataset. OSNet is chosen for its capability to capture discriminative features suitable for re-identification of athletes with similar appearances, which is critical for our tracklet refinement process.

\noindent\textbf{Hyperparameters.} 
We set minimum cluster samples $s$ to 5, maximum neighbor distance threshold $\epsilon$ to 0.6,  maximum clusters $k$ to 3, merging threshold $\alpha$ to 0.4, and $\beta$ to 1 for SportsMOT dataset and 0.7 for SoccerNet dataset, respectively.


\subsection{Performance}

\begin{table}[t]
\centering
\scriptsize
\caption{Tracking performance on SportsMOT before and after applying our Global Tracklet Association (GTA) method.}
\label{tab:results-sportsmot}
\begin{tabular}{l|ccccccc}
    \hline
    \hline
    Method & HOTA$\uparrow$ & AssA$\uparrow$ & IDF1$\uparrow$ & DetA$\uparrow$ & MOTA$\uparrow$ & IDs$\downarrow$ \\
    \hline
    SORT \cite{DeepSORT} & 56.28 & 42.67 & 58.83 & 74.30 & 85.11 & 5180 \\
    \rowcolor{lightgray}
    SORT + GTA & \textbf{66.52} \tiny\textcolor{OliveGreen}{($+$10.24)} & \textbf{59.59} \tiny\textcolor{OliveGreen}{($+$16.92)} & \textbf{77.37} \tiny\textcolor{OliveGreen}{($+$18.54)} & 74.29 & \textbf{85.27} & \textbf{3547} \tiny\textcolor{OliveGreen}{(-1633)}\\
    \hline
    ByteTrack \cite{ByteTrack} & 63.46 & 51.81 & 70.76 & 77.81 & 94.91 & 3147 \\
    \rowcolor{lightgray}
    ByteTrack + GTA & \textbf{69.74} \tiny\textcolor{OliveGreen}{($+$6.28)} & \textbf{62.61} \tiny\textcolor{OliveGreen}{($+$10.80)} & \textbf{83.16} \tiny\textcolor{OliveGreen}{($+$12.40)} & 77.72 & \textbf{95.01} & \textbf{2107} \tiny\textcolor{OliveGreen}{($-$1040)}\\
    \hline
    Deep-EIoU \cite{huang2024iterative} & 77.21 & 67.63 & 79.81 & 88.22 & 96.30 & 2909 \\
    \rowcolor{lightgray}
    Deep-EIoU + GTA & \textbf{81.04} \tiny\textcolor{OliveGreen}{($+$3.83)} & \textbf{74.51} \tiny\textcolor{OliveGreen}{($+$6.88)} & \textbf{86.51} \tiny\textcolor{OliveGreen}{($+$6.70)} & 88.21 & \textbf{96.32} & \textbf{2737} \tiny\textcolor{OliveGreen}{($-$172)}\\
    \hline
    \hline
\end{tabular}
\end{table}

\noindent{\textbf{Evaluation Metrics.}} We utilize commonly used tracking metrics, including HOTA \cite{luiten2021hota} for its comprehensive evaluation of both detection and association accuracy (DetA and AssA); and CLEAR metrics \cite{CLEARMOT}, where IDF1 and MOTA serve as the standard benchmark for tracking performance across various scenarios, and  IDs to verify the effectiveness of our method in reducing and associating the correct identities.

\noindent{\textbf{Performance on SportsMOT.}} In Table \ref{tab:results-sportsmot}, our proposed method demonstrates significant performance improvements over existing trackers across various tracking metrics like HOTA, AssA, IDF1, IDs, and Frag. For the SportsMOT dataset, GTA achieved the highest HOTA improvement of 10.24\% for SORT and 3.83\% for Deep-EIoU, reaching a state-of-the-art HOTA of 81.04\%. Demonstrating the GTA method is applicable for diverse kinds of sports tracking.

\begin{figure}[t]
    \centering
    \includegraphics[width=1\textwidth]{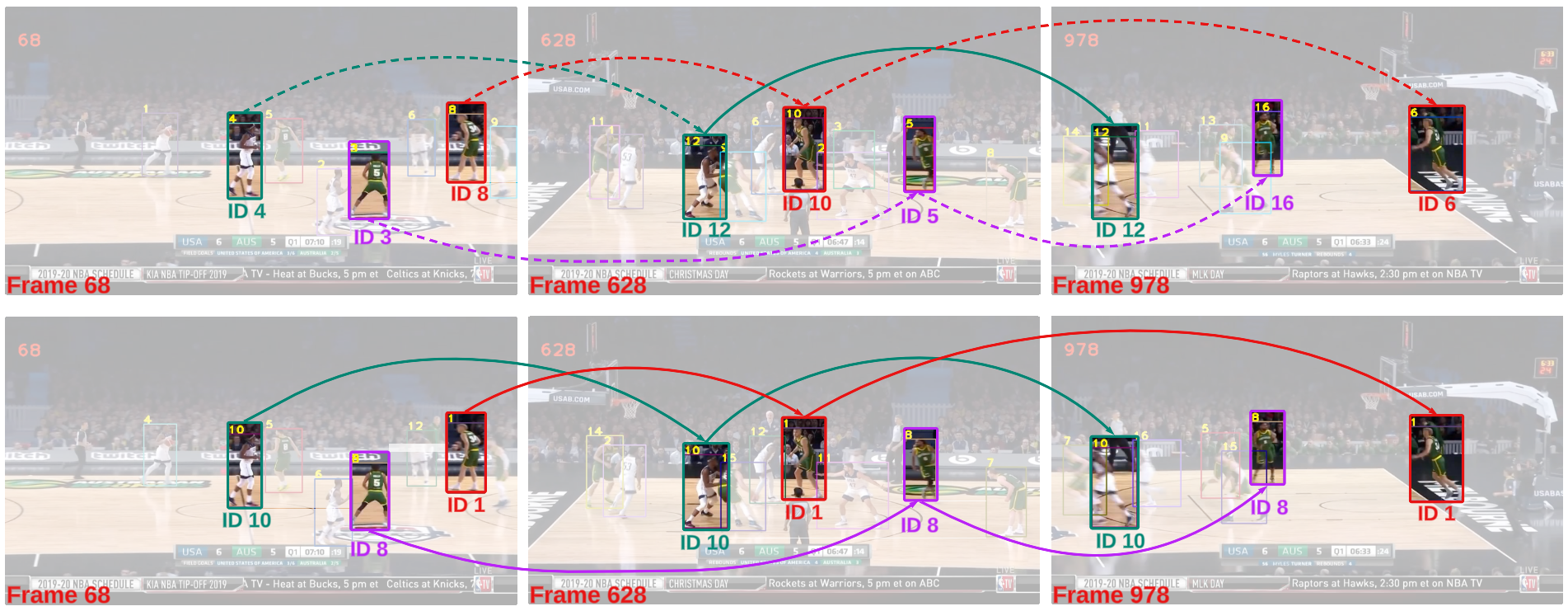}
    \includegraphics[width=1\textwidth]{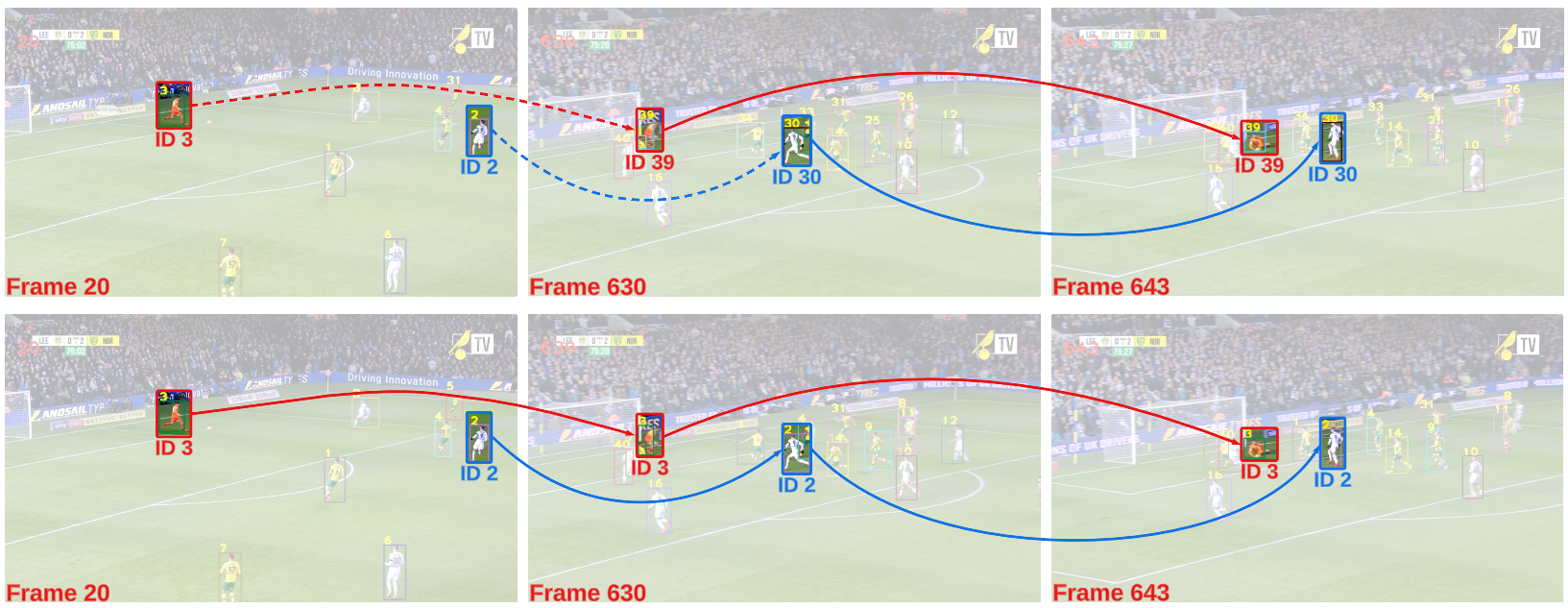}
    \vspace{-1em}
    \caption{Tracking visualization of athletes before and after applying Global Tracklet Association (GTA). In rows one and three, dashed lines indicate association errors, showing inconsistent athlete IDs across frames. In contrast, solid lines represent correct associations with consistent IDs after applying GTA (rows two and four). The comparison highlights how the algorithm improves ID continuity across frames in both basketball (first two rows) and soccer (second two rows) sequences.}
    \label{fig:before-after-frame-merge}
\end{figure}

\noindent{\textbf{Performance on SoccerNet.}} In Table \ref{tab:results-soccernet}, GTA improved HOTA by 6.84\% for SORT and 3.7\% for Deep-EIoU. These results demonstrate the effectiveness of our tracklet refinement method in enhancing tracker performance on challenging sports player tracking datasets. Figure \ref{fig:before-after-frame-merge} illustrates the qualitative results, showing GTA effectively connecting tracklet fragments from online tracking.

\begin{table}[t]
\centering
\scriptsize
\caption{Tracking performance on SoccerNet before and after applying our Global Tracklet Association (GTA) method.}
\label{tab:results-soccernet}
\begin{tabular}{l|ccccccc}
    \hline
    \hline
    Method & HOTA$\uparrow$ & AssA$\uparrow$ & IDF1$\uparrow$ & DetA$\uparrow$ & MOTA$\uparrow$ & IDs$\downarrow$\\
    \hline
    SORT \cite{DeepSORT} & 65.89 & 57.15 & 68.56 & 76.11 & 82.59 & 3281\\
    \rowcolor{lightgray}
    SORT + GTA & \textbf{72.73} \tiny\textcolor{OliveGreen}{(+6.84)} & \textbf{69.62} \tiny\textcolor{OliveGreen}{(+12.47)} & \textbf{81.24} \tiny\textcolor{OliveGreen}{(+12.68)} & 76.04 & \textbf{82.93} & \textbf{1374} \tiny\textcolor{OliveGreen}{(-1907)} \\
    \hline
    ByteTrack \cite{ByteTrack} & 67.30 & 60.38 & 73.22 & 75.14 & 84.66 & 4558\\
    \rowcolor{lightgray}
    ByteTrack + GTA & \textbf{71.97} \tiny\textcolor{OliveGreen}{(+4.67)} & \textbf{69.03} \tiny\textcolor{OliveGreen}{(+8.65)} & \textbf{82.67} \tiny\textcolor{OliveGreen}{(+9.45)} & 75.10 & \textbf{84.91} & \textbf{3149} \tiny\textcolor{OliveGreen}{(-1409)} \\
    \hline
    Deep-EIoU \cite{huang2024iterative} & 79.41 & 71.55 & 78.40 & 88.14 & 87.92 & 2803\\
    \rowcolor{lightgray}
    Deep-EIoU + GTA & \textbf{83.11} \tiny\textcolor{OliveGreen}{(+3.70)} & \textbf{78.38} \tiny\textcolor{OliveGreen}{(+6.83)} & \textbf{84.66} \tiny\textcolor{OliveGreen}{(+6.26)} & 88.13 & \textbf{88.03} & \textbf{2188} \tiny\textcolor{OliveGreen}{(-615)} \\
    \hline
    \hline
\end{tabular}
\end{table}

\subsection{Ablation Study}
To evaluate the effectiveness of the splitter and connector modules of our proposed method, we conduct ablation studies on the performance gain of SORT \cite{SORT}, ByteTrack\cite{ByteTrack}, and DeepEIoU\cite{huang2024iterative} after applying each module. The ablation study summarized in Table \ref{tab:effectiveness} highlights the effectiveness of different modules of the Global Tracklet Association (GTA) method on the performance of the three trackers across the SportsMOT and SoccerNet datasets. The proposed Connector alone results in notable improvements for SORT in HOTA, IDF1, and AssA scores, with an increase of 9.15\% in HOTA on SportsMOT and 5.85\% on SoccerNet. When the Splitter and Connector are both applied for SORT, we obtained HOTA improvements of 10.24\% on SportsMOT and 6.84\% on SoccerNet, along with substantial increases in IDF1 and AssA, demonstrating the effectiveness of both modules.

\begin{figure}[H]
    \centering
    \includegraphics[width=1\linewidth]{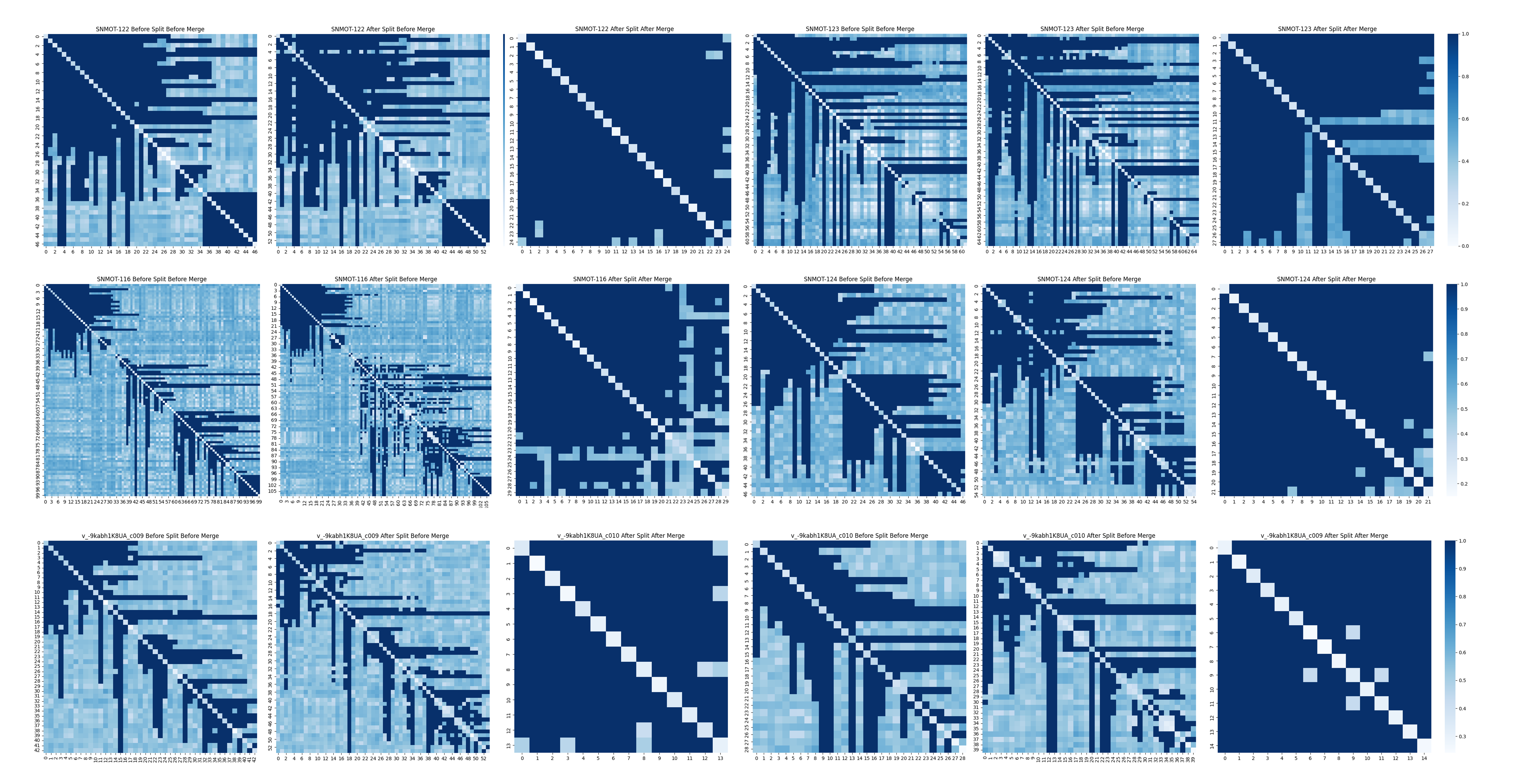}
    \caption{The cosine distance matrix of the embeddings of three stages: (1) Before Split, (2) After Split, and (3) After Connect. Both the x-axis and y-axis represent the IDs, while the darker color represents the farther distance.}
    \label{fig:enter-label}
\end{figure}

\begin{table}[t]
\centering
\scriptsize
\caption{Effectiveness of \textit{\textbf{Splitter}} and \textit{\textbf{Connector}} on different MOT algorithms and datasets.}
\label{tab:effectiveness}
\begin{tabular}{c|l|c c|c|c|c}
\hline
\hline
\textbf{Dataset} & \textbf{Method} & \textbf{\textit{Connector}} & \textbf{\textit{Splitter}} & \textbf{HOTA}$\uparrow$ & \textbf{AssA}$\uparrow$ & \textbf{IDF1}$\uparrow$\\
\hline
\multirow{9}{*}{\centering SportsMOT} & & & & 56.28 & 42.67 & 58.83\\
& SORT\cite{SORT} & \checkmark & & 65.43 \tiny\textcolor{OliveGreen}{(+9.15)} & 57.77 \tiny\textcolor{OliveGreen}{(+15.10)} & 76.13 \tiny\textcolor{OliveGreen}{(+17.30)}\\
& & \checkmark & \checkmark & 66.52 \tiny\textcolor{OliveGreen}{(+10.24)} & 59.59 \tiny\textcolor{OliveGreen}{(+16.92)} & 77.37 \tiny\textcolor{OliveGreen}{(+18.54)}\\
\cline{2-7}
& & & & 63.46 & 51.81 & 70.76 \\
& ByteTrack\cite{ByteTrack} & \checkmark & & 69.51 \tiny\textcolor{OliveGreen}{(+6.05)} & 62.21 \tiny\textcolor{OliveGreen}{(+10.40)} & 82.88 \tiny\textcolor{OliveGreen}{(+12.12)}\\
& & \checkmark & \checkmark & 69.74 \tiny\textcolor{OliveGreen}{(+6.28)} & 62.61 \tiny\textcolor{OliveGreen}{(+10.80)} & 83.16 \tiny\textcolor{OliveGreen}{(+12.40)}\\
\cline{2-7}
& & & & 77.21 & 67.63 & 79.81\\
& DeepEIoU\cite{huang2024iterative} & \checkmark & & 80.48 \tiny\textcolor{OliveGreen}{(+3.27)} & 73.50 \tiny\textcolor{OliveGreen}{(+5.87)} & 85.76 \tiny\textcolor{OliveGreen}{(+5.95)} \\
& & \checkmark & \checkmark & 81.04 \tiny\textcolor{OliveGreen}{(+3.83)} & 74.51 \tiny\textcolor{OliveGreen}{(+6.88)} & 86.51 \tiny\textcolor{OliveGreen}{(+6.70)} \\
\hline
\multirow{9}{*}{\centering SoccerNet} & & & & 65.89 & 57.15 & 68.56\\
& SORT & \checkmark & & 71.74 \tiny\textcolor{OliveGreen}{(+5.85)} & 67.74 \tiny\textcolor{OliveGreen}{(+10.59)} & 79.73 \tiny\textcolor{OliveGreen}{(+11.17)}\\
& & \checkmark & \checkmark & 72.73 \tiny\textcolor{OliveGreen}{(+6.84)} & 69.62 \tiny\textcolor{OliveGreen}{(+12.47)} & 81.24 \tiny\textcolor{OliveGreen}{(+12.68)}\\
\cline{2-7}
& & & & 67.30 & 60.38 & 73.22\\
& ByteTrack & \checkmark & & 71.05 \tiny\textcolor{OliveGreen}{(+3.75)} & 67.30 \tiny\textcolor{OliveGreen}{(+6.92)} & 78.22 \tiny\textcolor{OliveGreen}{(+4.61)}\\
& & \checkmark & \checkmark & 71.97 \tiny\textcolor{OliveGreen}{(+4.67)} & 69.03 \tiny\textcolor{OliveGreen}{(+8.65)} & 82.67 \tiny\textcolor{OliveGreen}{(+9.45)}\\
\cline{2-7}
& & & & 79.41 & 71.55 & 78.40\\
& DeepEIoU & \checkmark & & 82.01 \tiny\textcolor{OliveGreen}{(+2.60)} & 76.32 \tiny\textcolor{OliveGreen}{(+4.77)} & 83.13 \tiny\textcolor{OliveGreen}{(+4.73)}\\
& & \checkmark & \checkmark & 83.11 \tiny\textcolor{OliveGreen}{(+3.70)} & 78.38 \tiny\textcolor{OliveGreen}{(+6.83)} & 84.66 \tiny\textcolor{OliveGreen}{(+6.26)}\\
\hline
\hline
\end{tabular}
\end{table}

\section{Conclusion}
In this paper, we proposed the Global Tracklet Association (GTA), a novel tracklet refinement method to enhance the performance of existing trackers in challenging sports player tracking scenarios. Our approach effectively addresses common issues such as mix-up errors and cut-off errors by leveraging a combination of ReID model and unsupervised clustering techniques for tracklet splitting and merging. The integration of our GTA method with trackers like SORT, ByteTrack, and Deep-EIoU has demonstrated significant improvements in various tracking performance metrics, particularly in metrics that related to associations like HOTA, AssA, and IDF1, while also reducing the number of ID switches (IDs) and tracklet fragments (Frag). Our proposed module achieves state-of-the-art performance on SportsMOT and SoccerNet datasets. Future work will focus on exploring additional enhancements and adaptations of the GTA method for other multi-object tracking scenarios beyond sports.

%
%
\bibliographystyle{splncs04}
\bibliography{main}
\end{document}